\documentclass{article}

\usepackage[preprint]{neurips_2026}        

\usepackage[utf8]{inputenc}
\usepackage[T1]{fontenc}
\usepackage{hyperref}
\usepackage{url}
\usepackage{booktabs}
\usepackage{amsfonts}
\usepackage{amsmath}
\usepackage{nicefrac}
\usepackage{microtype}
\usepackage{xcolor}
\usepackage{graphicx}
\usepackage{amssymb}
\usepackage{algorithm}
\usepackage{algpseudocode}
\usepackage{minted}
\usepackage{listings}
\usepackage{xcolor}
\usepackage{multirow}
\usepackage{makecell}
\usepackage{subcaption}

\definecolor{codeblue}{rgb}{0.25,0.5,0.5}
\definecolor{codekw}{rgb}{0.85, 0.18, 0.50}

\definecolor{codesign}{RGB}{0, 0, 255}
\definecolor{codefunc}{rgb}{0.85, 0.18, 0.50}

\lstdefinelanguage{PythonFuncColor}{
language=Python,
keywordstyle=\color{blue}\bfseries,
commentstyle=\color{codeblue},
stringstyle=\color{orange},
showstringspaces=false,
basicstyle=\ttfamily\small,
literate=
  {*}{{\color{codesign}* }}{1}
  {-}{{\color{codesign}- }}{1}
  {+}{{\color{codesign}+ }}{1}
  {/}{{\color{codesign}/ }}{1}
  {(}{{\color{codesign}(}}{1}
  {)}{{\color{codesign})}}{1}
  {model}{{\color{codefunc}model}}{1}
  {velocity_to_x0}{{\color{codefunc}velocity\_to\_x0}}{1}
  {perceptual_loss}{{\color{codefunc}perceptual\_loss}}{1}
  {decode}{{\color{codefunc}decode}}{1}
  {stopgrad}{{\color{codefunc}stopgrad}}{1}
}

\title{Perceptual Flow Matching for Few-Step Generative Modeling}

\author{%
Chuyang Zhao$^{1}$\thanks{Contact: \texttt{zhaochuyang.3@jd.com}}
\quad Yifei Song$^{2}$
\quad Hongfa Wang$^{3}$
\quad Jianlong Yuan$^{1}$ \\
Yuan Zhang$^{1}$
\quad Siming Fu$^{1}$
\quad Zhineng Chen$^{2}$
\quad Huilin Deng$^{4}$
\quad Haoyang Huang$^{1}$
\quad Nan Duan$^{1}$\thanks{Corresponding author.} \\[0.5em]
$^{1}$Joy Future Academy \quad
$^{2}$Fudan University \quad
$^{3}$Tsinghua University \quad
$^{4}$USTC \\
}

\begin{document}

\maketitle

\begin{abstract}

We propose \textbf{Perceptual Flow Matching} (\textbf{PFM}), a simple yet effective framework for few-step generation in flow-matching models. Rather than performing velocity regression in the conventional VAE latent space, PFM supervises flow matching in a \emph{perceptual feature space} using pretrained perceptual models. This simple change substantially improves the few-step generation capability of flow-matching models, reducing the number of sampling steps from $35$--$50$ to $4$--$8$ while preserving generation quality. Unlike existing acceleration and distillation approaches, PFM requires neither teacher models nor auxiliary score networks and can be integrated into standard flow-matching training pipelines with minimal modifications. Extensive experiments on image generation, video generation, and image editing tasks demonstrate that PFM consistently produces high-quality results while producing fewer artifacts than existing distillation-based methods. 
We further show that perceptual supervision shifts the regression minimizer from mean-seeking to mode-seeking, biasing predictions toward on-manifold modes that remain accurate under coarse few-step integration.
Our results reveal that standard flow-matching training can naturally yield high-quality few-step generators when supervised in an appropriate representation space. We hope this insight inspires future research into representation-aware objectives for efficient generative modeling.

\end{abstract}

\section{Introduction}

Flow matching has emerged as a powerful framework for generative modeling and has achieved remarkable success across image generation, image editing, and video generation~\citep{lipman2023flow, labs2025flux, esser2024scaling,liu2023rectifiedflow,wan2025wan,wu2025qwenimage, bai2026mainecoon}. However, despite its impressive generation quality, practical deployment remains challenging due to the high computational cost of inference. Similar to diffusion models, flow matching models typically require dozens of sampling steps (e.g., 35--50 steps) to produce high-quality outputs, resulting in substantial latency and computational overhead.

To address this challenge, prior works have explored few-step generation through distillation~\citep{salimans2022progressive,yin2024dmd,yin2024dmd2,dao2025scflow}, consistency training~\citep{song2023consistency}, and continuous-time flow field modeling~\citep{lu2025simplifying,geng2025meanflow,sabour2025alignyourflow}. 
While these approaches have demonstrated promising results, they often require auxiliary models, or substantial modifications to the original training framework, increasing both computational cost and implementation complexity.
Moreover, distillation-based approaches can be prone to overfitting and mode collapse, which leads to undesirable synthetic artifacts~\citep{geng2025meanflow}.

We propose Perceptual Flow Matching (PFM), a simple yet effective framework for few-step generation. Rather than supervising flow matching through conventional velocity regression in the VAE latent space, PFM formulates the training objective as a perceptual regression loss on decoded samples, using feature representations extracted from pretrained perceptual models~\citep{zhang2018lpips,oquab2023dinov2}.
We further show that classifier-free guidance can be naturally incorporated during training~\citep{ho2022classifierfree}, eliminating the need for an additional unconditional forward pass at inference time. 
Notably, PFM requires no teacher models, auxiliary score networks, or distillation procedures. It retains the original flow matching framework~\citep{lipman2023flow} and only replaces the supervision space, making it easy to implement and reducing training cost.
As PFM is trained directly on real data and closely follows the standard flow matching training paradigm, requiring only a change in the supervision space, it is scalable to large-scale training and produces fewer synthetic artifacts compared to distillation-based methods.

The key insight we find is that the supervision space determines the geometry of the one-step prediction. Under the standard Euclidean objective, flow matching is effectively mean-seeking: when the posterior $q(x_0\mid x_t)$ is multimodal at high noise levels, the optimal prediction lies between plausible data modes, producing an off-manifold and blurry estimate that is difficult to correct with only a few sampling steps. In contrast, pretrained perceptual models encode the structure of natural images in perceptual space, assigning larger distances to off-manifold degradations such as blur while keeping valid images in coherent manifold. As a result, perceptual supervision makes blurry averages less favorable and biases the prediction toward perceptually valid samples, substantially improving few-step generation.

We conduct extensive experiments on image generation, image editing, and video generation tasks to validate the effectiveness of PFM. 
On the text-to-image COCO 2014 val benchmark, PFM achieves state-of-the-art performance among 8-step generation methods, attaining an FID of 33.93, a CLIP score of 31.70 and a HPSv3 score of 11.42.
On image editing and video generation benchmarks, PFM substantially outperforms its few-step baseline counterparts, while achieving results comparable to, and in some cases surpassing, the original multi-step models.
Our results suggest that standard flow-matching training can naturally yield strong few-step generators when optimized in an appropriate representation space. 
We hope this insight motivates further investigation into the role of supervision space in generative modeling.

\section{Related Work}

\subsection{Few-Step Generation and Distillation}

Reducing the sampling cost of diffusion and flow-based generative models has been a central research topic in recent years. Existing approaches can largely be categorized into four paradigms. Progressive Distillation~\citep{salimans2022progressive} accelerates generation by repeatedly training a student model to emulate multiple sampling steps of a pretrained teacher, gradually reducing the number of required inference steps. Consistency Models and Consistency Distillation~\citep{song2023consistency} learn a trajectory-consistent mapping that enables one-step or few-step generation through teacher-generated trajectories and consistency constraints. Distribution Matching Distillation (DMD)~\citep{yin2024dmd} and its variants \citep{yin2024dmd2, yin2025slow} align the distribution of student-generated samples with those of a stronger teacher through additional score estimators, critics, or auxiliary objectives. More recently, methods such as MeanFlow~\citep{geng2025meanflow} have explored alternative training targets for flow matching, replacing the conventional instantaneous velocity objective with an average-velocity formulation and enforcing consistency between the two through Jacobian-vector product (JVP).

\subsection{Perceptual Supervision}

Perceptual losses~\citep{johnson2016perceptual, zhang2018lpips} have been widely adopted in image synthesis, image restoration, and representation learning to improve perceptual quality. Existing works~\citep{yu2024representation, leng2025repa, wu2026representation} primarily employ perceptual supervision to improve reconstruction fidelity or visual realism. In contrast, we investigate perceptual supervision as a mechanism for enabling few-step generation. Our results suggest that the invariances encoded by perceptual representations fundamentally improve the few-step capability of flow matching models, eliminating the need for explicit distillation.

\section{Method}
\subsection{Flow Matching Preliminaries}

Flow Matching (FM)~\citep{lipman2023flow} learns a continuous transport map between a simple noise distribution and the data distribution. Given a clean latent representation $x_0$ and Gaussian noise $\epsilon \sim \mathcal{N}(0,I)$, an interpolated sample is constructed as:
\begin{equation}
x_t = (1-\sigma_t)x_0 + \sigma_t \epsilon,
\label{eq:interpolation}
\end{equation}
where $\sigma_t \in [0,1]$ denotes the noise level associated with timestep $t$. The corresponding target velocity field is:
\begin{equation}
v(x_t,t) = \epsilon - x_0.
\label{eq:target_velocity}
\end{equation}

A flow matching model $v_\theta(x_t,t,c)$ is trained to predict this velocity field by minimizing a mean squared error objective:
\begin{equation}
\mathcal{L}_{\mathrm{FM}}
=
\mathbb{E}_{x_0,\epsilon,t}
\left[
\left\|
v_\theta(x_t,t,c)
-
v(x_t,t)
\right\|_2^2
\right],
\label{eq:fm_loss}
\end{equation}
where $c$ denotes optional conditioning information such as text prompts or reference images. Given a predicted velocity field, a clean sample estimate can be recovered by inverting from Eq.~\ref{eq:interpolation} as:
\begin{equation}
\hat{x}_0 = x_t - \sigma_t v_\theta(x_t,t,c).
\label{eq:x0_prediction}
\end{equation}

Conventional flow matching applies supervision directly to the velocity prediction in Eq.~\ref{eq:fm_loss}. In contrast, our method revisits the supervision space used during training and instead supervises the recovered clean prediction $\hat{x}_0$ in a perceptual feature space.

\subsection{Perceptual Flow Matching}

Conventional flow matching supervises the predicted velocity field through a regression objective in VAE latent space. In this work, we revisit the supervision space used during training and propose \textbf{Perceptual Flow Matching (PFM)}, which decodes the recovered latent prediction into pixel space and applies supervision in a perceptual feature space.

Given a noisy latent $x_t$, the model predicts a velocity field $v_\theta(x_t,t,c)$ and recovers a clean sample estimate $\hat{x}_0$ via Eq.~\ref{eq:x0_prediction}. Both the prediction $\hat{x}_0$ and the ground-truth sample $x_0$ are then decoded into pixel space using a pretrained VAE decoder $D(\cdot)$:
\begin{equation}
\hat{y}_0 = D(\hat{x}_0),
\qquad
y_0 = D(x_0).
\label{eq:decode}
\end{equation}

Instead of comparing $\hat x_0$ and $x_0$ in latent space, PFM measures the discrepancy of their decoded images in a perceptual feature space induced by $\phi(\cdot)$:
\begin{equation}
\mathcal{L}_{\mathrm{PFM}} = \mathbb{E}_{x_0,\epsilon,t}\left[ d\bigl(\phi(\hat y_0),\phi(y_0)\bigr) \right],
\label{eq:pfm_loss}
\end{equation}
where $d(\cdot,\cdot)$ denotes a perceptual distance metric.

At inference time, PFM models can be sampled using standard consistency
sampling~\citep{song2023consistency,yin2024dmd}: at each step, the model predicts $\hat{x}_0$ from $x_t$ via
Eq.~\ref{eq:x0_prediction}, which is then re-noised to obtain $x_{t'}$
for the next timestep. The number of sampling steps can be freely
adjusted without retraining, allowing users to trade off quality and
speed as needed.

\subsection{Classifier-free Guidance}
\label{sec:cfg}

Classifier-free guidance (CFG)~\citep{ho2022classifierfree} is a widely adopted inference-time technique for improving generation quality and condition alignment. Given conditional and unconditional predictions, CFG constructs a guidance-enhanced prediction as:
\begin{equation}
v_{\mathrm{cfg}}
=
v_\theta(x_t,t,\varnothing)
+
w \Bigl(
v_\theta(x_t,t,c)
-
v_\theta(x_t,t,\varnothing)
\Bigr),
\end{equation}
where $w$ denotes the guidance scale. While effective, CFG requires an additional unconditional network evaluation at every sampling step, which increases inference cost. We explore two strategies to bake classifier-free guidance in training:

\textbf{Prediction-side.} Inspired by GFT~\citep{chen2025visual}, we incorporate guidance directly into the model's prediction before computing the perceptual loss:
\begin{equation}
\tilde{v}
= \beta \, v_\theta(x_t,t,c)
+ (1-\beta) \, \mathrm{sg}\bigl[v_\theta(x_t,t,\varnothing)\bigr],
\quad \beta = 1/w,
\label{eq:cfg_pred}
\end{equation}
where $\mathrm{sg}[\cdot]$ denotes the stop-gradient operator and $v_\theta(x_t,t,\varnothing)$ denotes the unconditional prediction. The perceptual loss is then applied to $\hat{x}_0$ recovered from $\tilde{v}$ via Eq.~\ref{eq:x0_prediction}, so that gradients flow through the conditional branch while the model learns to internalize the effect of guidance.

\textbf{Target-side.} Alternatively, guidance can be baked into the supervision target rather than the model's prediction. We construct a CFG-enhanced target and directly optimize the model toward it:
\begin{equation}
x_{\mathrm{cfg}}
=
\alpha x_0
+
(1-\alpha)
\Bigl[
\hat{x}_{0,\varnothing}
+
w
\bigl(
\hat{x}_{0,c}
-
\hat{x}_{0,\varnothing}
\bigr)
\Bigr],
\label{eq:cfg_target}
\end{equation}
where $\hat{x}_{0,c}$ and $\hat{x}_{0,\varnothing}$ denote the clean-sample estimates reconstructed from the conditional and unconditional predictions, respectively. $\alpha \in [0,1]$ controls the interpolation between the ground-truth sample and the CFG-enhanced target for training stability. 
The perceptual loss (Eq.~\ref{eq:pfm_loss}) is then computed using $x_{\mathrm{cfg}}$ as the supervision target, enabling guidance-enhanced generation in a single forward pass at inference.

Both strategies can be incorporated into the standard flow matching framework with minimal modification, requiring only an additional unconditional forward pass during training. Empirically, however, we find that PFM already substantially reduces the need for classifier-free guidance at inference time (Section~\ref{sec:abl_cfg}). Models trained solely with the perceptual objective can generate satisfactory results without classifier-free guidance. We attribute this to the learned perceptual supervision geometry, which biases the one-step prediction toward semantically valid samples. CFG baking can be optionally used to further strengthen the guidance effects.

\definecolor{codeblue}{rgb}{0.25,0.5,0.5}
\definecolor{codekw}{rgb}{0.85, 0.18, 0.50}

\definecolor{codesign}{RGB}{0, 0, 255}
\definecolor{codefunc}{rgb}{0.85, 0.18, 0.50}

\begin{algorithm}[t]
\caption{Perceptual Flow Matching (PFM) with optional CFG baking}
\label{alg:pfm}
\begin{lstlisting}
# ==========================================
# PFM
# ==========================================
v_pred = model(x_t, t, c)
x0_pred = velocity_to_x0(v_pred, x_t, t)

loss = perceptual_loss(
  decode(x0_pred),
  decode(x0)
)

# ==========================================
# PFM with prediction side CFG baking
# ==========================================
v_pred = model(x_t, t, c)
v_uncond = model(x_t, t, null)
v_pred = beta * v_pred + (1 - beta) * stopgrad(v_uncond)
x0_pred = velocity_to_x0(v_pred, x_t, t)

loss = perceptual_loss(
  decode(x0_pred),
  decode(x0)
)
\end{lstlisting}
\end{algorithm}

\subsection{Why Perceptual Space Works?}
\label{sec:why_perceptual}

\begin{figure}[t]
\centering
\hspace*{-0.05\linewidth}
\includegraphics[width=1.1\linewidth]{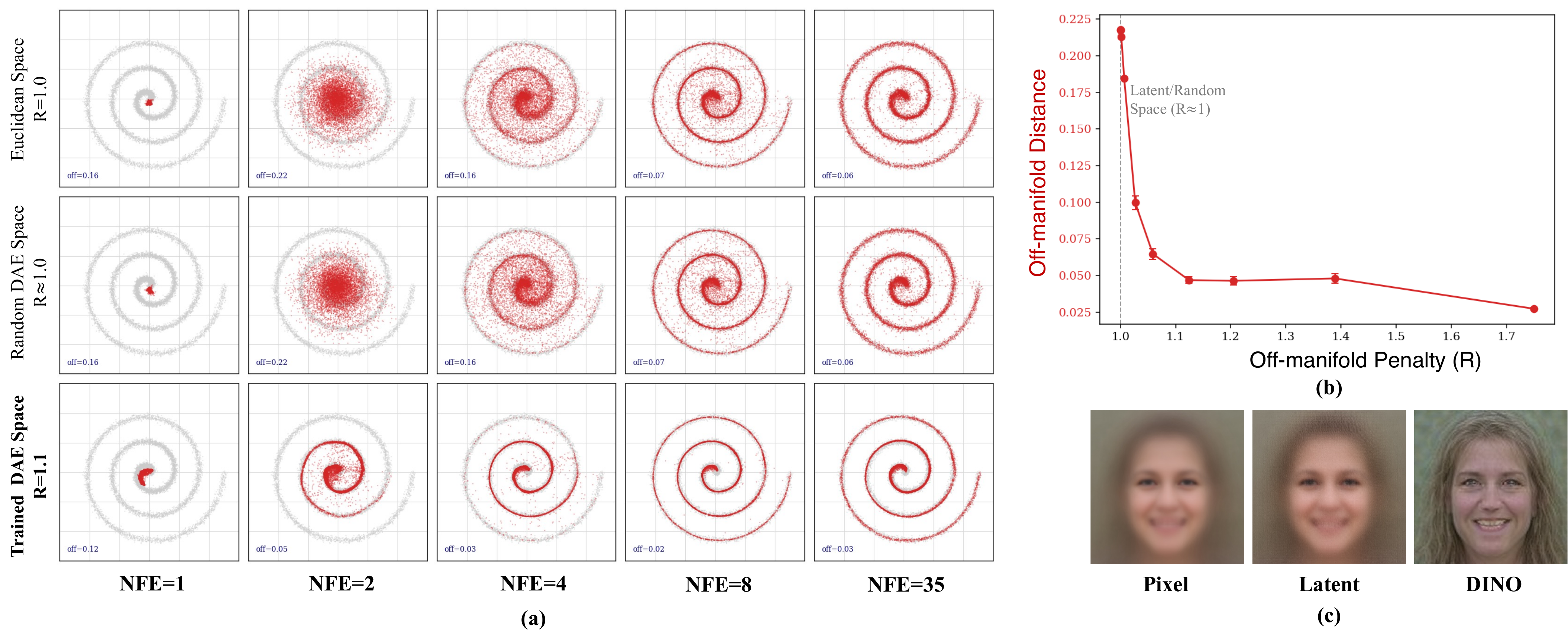}
\caption{
\textbf{(a)} Flow matching (FM) on 2D spiral data, supervised in three spaces: (i) velocity regression in Euclidean space (standard FM), (ii) $x_0$ regression in a random denoising-autoencoder (DAE) perceptual space, and (iii) $x_0$ regression in a pretrained DAE perceptual space. In Euclidean space, the prediction collapses toward the data mean at small NFE, whereas supervision in the pretrained DAE space places samples on the real data manifold even at small NFE. 
\textbf{(b)} Off-manifold distance (at NFE${=}2$) as the off-manifold penalty $R_\phi$ increases: larger $R_\phi$ yields lower off-manifold distance, i.e.\ better few-step generation.
\textbf{(c)} Average of 1{,}000 similar face images from FFHQ in pixel, VAE latent, and DINO feature space. The pixel and VAE-latent averages are blurry (off-manifold), whereas averaging in DINO feature space preserves sharp facial structure.
}
\label{fig:exp_toy_A}
\end{figure}

During training, standard flow matching minimizes the squared Euclidean distance between the model's one-step prediction of $\hat x_0$ and the true $x_0$\footnote{Standard flow matching predicts velocity $v$, an affine transform of $x_0$ ($\hat x_0=x_t-t v$ on the linear path), so the two objectives are equivalent and share the minimizer $\hat
  x_0^\star=\mathbb{E}[x_0\mid x_t]$.}; the minimizer of this prediction is the posterior mean
$\hat x_0^\star=\mathbb{E}[x_0\mid x_t]$. At high noise levels the posterior $q(x_0\mid x_t)$ is multimodal, so this mean may lie between data modes, producing an off-manifold and visually blurred prediction. 
Many-step sampling can gradually correct such estimates through iterative evaluations, but in the few-step regime there are too few updates to fully recover from this initial off-manifold prediction.

Perceptual supervision changes the geometry of the minimizer. Under the perceptual space, the minimizer becomes a feature-space barycenter rather than the Euclidean posterior mean. Because $\phi$ is pretrained on large-scale natural-image data, its representation encodes the structure of valid images: natural samples tend to lie in coherent feature manifold, whereas off-manifold artifacts such as blurry interpolations are assigned disproportionately large distances. As a result, the inter-mode average is no longer a cheap solution, and the learned prediction is biased away from blurry barycenters and toward perceptually valid samples that can be realized with few sampling steps. In this sense, perceptual supervision shifts the effective behavior of regression from Euclidean mean-seeking toward perceptual mode-seeking.

  \begin{table}[t]
    \renewcommand{\arraystretch}{1.25}
    \centering
    \begin{minipage}[t]{0.46\linewidth}
      \centering
      \captionof{table}{Off-manifold penalty $R_\phi$ across supervision spaces. \emph{Diff} is $R_{\phi}$ relative to the pixel space.}
      \label{tab:why_perceptual_spaces}
      \setlength{\tabcolsep}{3pt}
  \begin{tabular}{lcc}
  \toprule
  Supervision space & $R_{\phi}$ & Diff \\
  \midrule
  VAE Latent (standard FM)    & 1.594          & 0.000 \\
  VGG      & 1.946          & $+0.352$ \\
  DINO     & 1.940          & $+0.346$ \\
  SigLIP   & 1.900          & $+0.306$ \\
  ConvNeXt & \textbf{2.026} & $\mathbf{+0.432}$ \\
  CLIP     & 1.953          & $+0.359$ \\
  \bottomrule
  \end{tabular}
    \end{minipage}
    \hfill
    \begin{minipage}[t]{0.52\linewidth}
      \centering
      \captionof{table}{Effect of the perceptual-space dimension $D$ in the toy experiment. With the trained DAE feature space, increasing feature capacity raises $R_\phi$ and improves few-step performance.}
      \label{tab:why_perceptual_dim}
      \setlength{\tabcolsep}{5pt}
      \begin{tabular}{lccc}
      \toprule
      $D$ & $R_\phi$ & off-dist@NFE2 $\downarrow$ & off-dist@NFE4 $\downarrow$ \\
      \midrule
      2   & 1.095 & 0.116 & 0.075 \\
      8   & 1.165 & 0.068 & 0.036 \\
      16  & 1.187 & 0.054 & 0.029 \\
      64  & 1.205 & \textbf{0.046} & \textbf{0.027} \\
      \bottomrule
      \end{tabular}
    \end{minipage}
  \end{table}

To validate this mechanism, we measure how a supervision space treats off-manifold averages. Let $a$ and $b$ be two on-manifold samples, and let $m=\tfrac12(a+b)$ denote their Euclidean midpoint.
Although $a$ and $b$ are valid samples, their midpoint can lie off the data manifold, making $m$ a simple proxy for the blurry, averaged image produced by an $\ell_2$ regression objective.
Writing the feature-space distance as $d_\phi(u,v)=\lVert\phi(u)-\phi(v)\rVert$\footnote{Euclidean distance is the special case $\phi=\mathrm{identity}$.}, we define:
\begin{equation}
    R_\phi(a,b)=\frac{2 d_\phi(m,a)}{d_\phi(b,a)}.
    \label{eq:R}
\end{equation}
Since $m$ is the geometric midpoint of the chord $a-b$, any isometric or locally linear feature space gives $R_\phi\approx 1$.
In this case, the off-manifold midpoint is treated as a cheap interpolation.
In contrast, $R_\phi>1$ indicates that $\phi$ expands the distance from the midpoint to the data manifold, penalizing blurry averages and pushing the barycenter toward a valid mode.

Following the manifold assumption, we conduct a toy experiment. The underlying data lie on a smooth one-dimensional manifold (an Archimedean spiral) in a $\mathbb{R}^2$ latent space. 
We train a 5-layer ReLU MLP with 256-dim hidden units as the generator following~\citet{li2025backtobasics} on the 2D latent space. 
The perceptual model is a self-supervised denoiser autoencoder (DAE) trained on perceptual space $\mathbb{R}^D$, where the 2D latent is lifted by a fixed column-orthogonal projection $P\in\mathbb{R}^{D\times d}$ ($P^\top P=I_d$).
We compare three supervision spaces: (1) standard velocity regression in the latent space, (2) regression through the feature space of an untrained DAE, and (3) regression through the feature space of a trained DAE.
As shown in Figure~\ref{fig:exp_toy_A}, both latent-space supervision and the untrained-DAE feature space exhibit mean-seeking behavior: with few sampling steps, their samples concentrate near the off-manifold average of the distribution, the toy analog of pixel-space blur. These samples gradually move back onto the manifold only when many integration steps are used. In contrast, supervision through the trained DAE feature space, which yields ($R_\phi>1$), produces samples that lie close to the manifold with far fewer steps.
The untrained DAE provides a nonlinear, non-Euclidean feature space, yet its $R_\phi\approx 1$ and its failure indicate that the key factor is the learned off-manifold penalty rather than feature nonlinearity alone; this is consistent with our RandViT ablation on SD3-Medium (Section~\ref{sec:abl_pm}).
As shown in Figure~\ref{fig:exp_toy_A}, few-step quality improves as the off-manifold penalty $R_\phi$ increases: larger $R_\phi$ consistently reduces the off-manifold distance of generated samples and improves few-step generation performance.  Table~\ref{tab:why_perceptual_dim} further shows that increasing the feature dimension $D$ of the perceptual DAE space raises $R_\phi$ and improves few-step generation performance.

Finally, we verify that pretrained perceptual spaces exhibit a stronger off-manifold penalty on natural images. For each anchor image $a$ from FFHQ~\citep{karras2019style}, we retrieve its $k$ nearest neighbors in pixel space ($k{=}8$) and select one neighbor as another on-manifold sample $b$. Since real images do not provide an exact inter-mode midpoint, we construct a practical off-manifold proxy $m$ by averaging the 100 images most similar to $a$. This averaged face is visibly blurry and off-manifold, resembling the blur commonly produced by coarse few-step prediction. We then measure $R_\phi$ in the SD3 VAE latent space and in several pretrained perceptual spaces (Table~\ref{tab:why_perceptual_spaces}).
The VAE latent space gives the weakest penalty ($R_\phi\approx 1.59$), treating the blurry average as only moderately farther from the anchor than another real image. Thus, under coarse one-step prediction, latent-space supervision provides limited pressure against blur, helping explain why few-step sampling struggles even though many-step integration can still recover high-quality samples. In contrast, pretrained perceptual features such as VGG, DINOv2, and ConvNeXt assign substantially larger penalties ($R_\phi\approx 1.9$--$2.0$).
Surprisingly, the ranking of $R_\phi$ across supervision spaces closely matches their relative performance on the text-to-image benchmark (Table~\ref{tab:backbone_comparison}).
This indicates $R_\phi$ plays a vital role in the few-step generation performance.

Furthermore, we examine why averaging in perceptual space is more compatible with few-step generation. We take $1{,}000$ similar face images from FFHQ and average them in three spaces: (1) directly in pixel space, (2) in SD3 VAE latent space, decoded with the SD3 VAE decoder, and (3) in DINOv2 feature space, decoded with a representation autoencoder (RAE)~\citep{zheng2025rae}. As shown in Figure~\ref{fig:exp_toy_A}(c),
the DINOv2 average preserves sharp facial structure, whereas the pixel and latent averages are blurry.
This suggests that averaging in a semantic feature space is far more benign than averaging in pixel or latent space. Because pretrained perceptual features organize images along semantically meaningful dimensions, their average is more likely to decode to a coherent on-manifold image rather than an off-manifold blur. As a result, even when the high-noise posterior is multimodal, perceptual supervision provides a more coherent one-step target, improving robustness under coarse few-step integration.

  \begin{table}[t]
    \setlength{\tabcolsep}{15pt}
    \renewcommand{\arraystretch}{1.25}
    \caption{\textbf{Text-to-image generation results on COCO 2014 val set.} We manually implemented LCM and DMD2 on SD3-Medium based on their original codebase and trained on the same dataset.}
    \label{tab:t2i_results}
    \centering
    \begin{tabular}{lcccc}
    \toprule
    Method & NFE & FID $\downarrow$ & CLIP $\uparrow$ & HPSv3 $\uparrow$\\
    \midrule
    SD3-Medium   & $50\times 2$ & 28.26 & 33.5384 & 10.90 \\
    \midrule
    LCM          & 4 & \textbf{30.59} & 30.62 & 6.68 \\
    DMD2         & 4 & 36.20 & 31.24 & 8.28 \\
    PFM (Ours)   & 4 & 31.16 & \textbf{31.89} & \textbf{8.50} \\
    \midrule
    LCM          & 8 & 34.18 & 30.42 & 9.19  \\
    DMD2         & 8 & 36.16 & 31.54 & 10.19 \\
    PFM (Ours)   & 8 & \textbf{33.93} & \textbf{31.70} & \textbf{11.42}  \\
    \bottomrule
    \end{tabular}
  \end{table}
  
  \begin{figure}[t]
    \centering
    \begin{subfigure}[t]{0.48\linewidth}
      \centering
      \includegraphics[width=\linewidth]{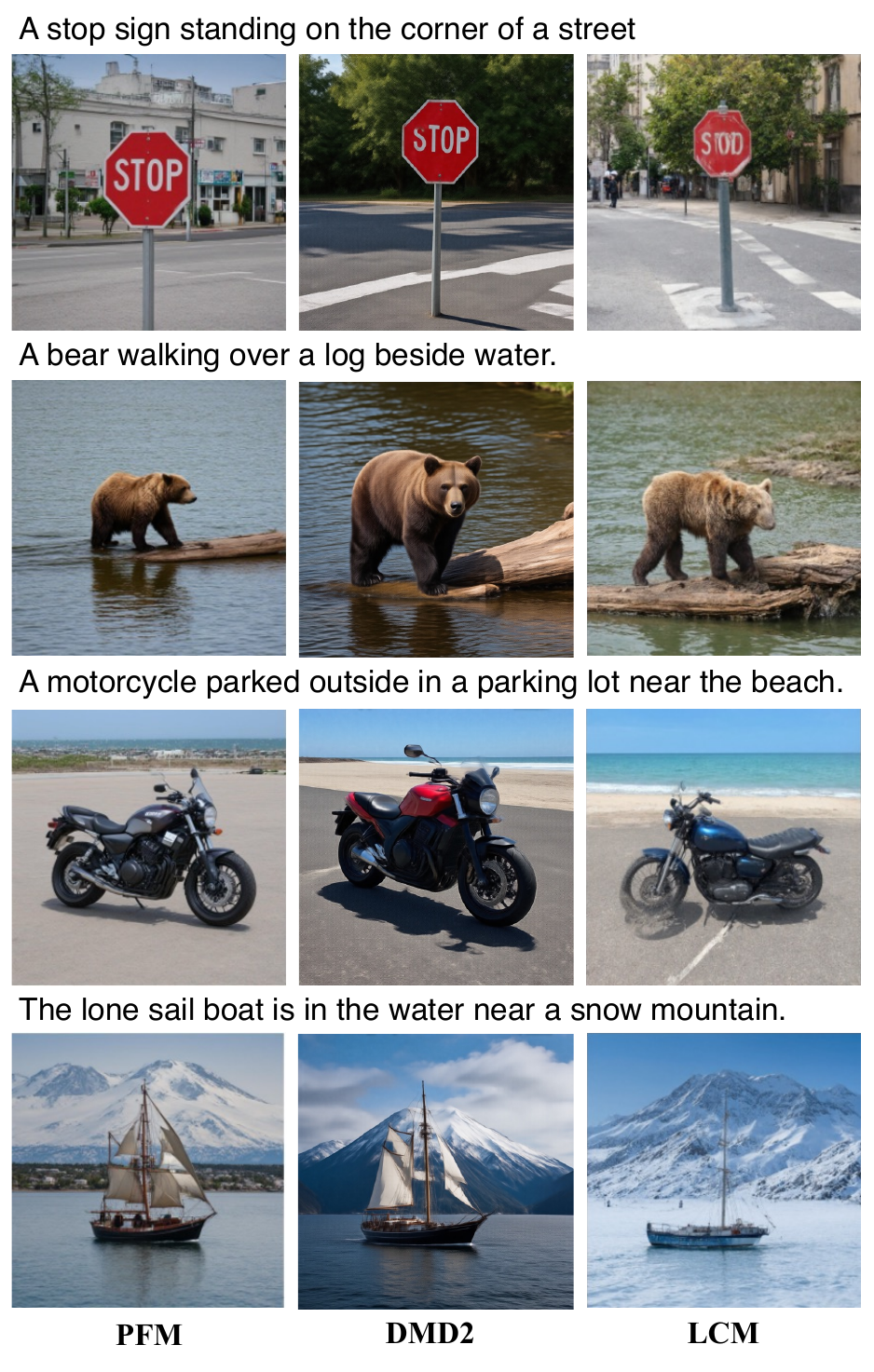}
      \caption{NFE=4 comparison}
      \label{fig:exp_sd3_nfe4}
    \end{subfigure}
    \hfill
    \hspace{-2mm}
    \begin{subfigure}[t]{0.495\linewidth}
      \centering
      \includegraphics[width=\linewidth]{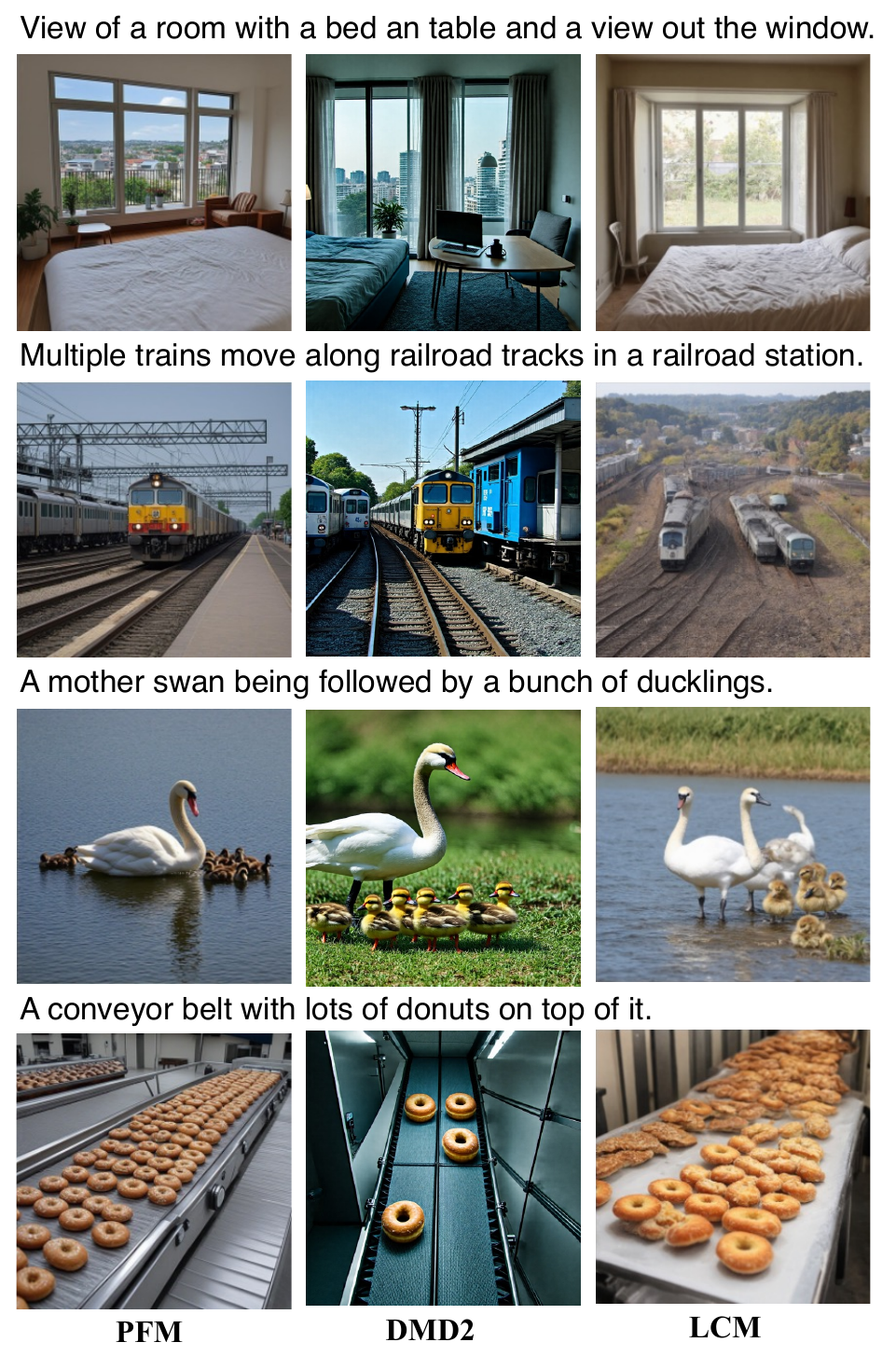}
      \caption{NFE=8 comparison}
      \label{fig:exp_sd3_nfe8}
    \end{subfigure}
    \caption{Qualitative comparison on SD3-Medium at different sampling steps.}
    \label{fig:exp_sd3}
  \end{figure}

\begin{figure}[t]
\centering
\includegraphics[width=\linewidth]{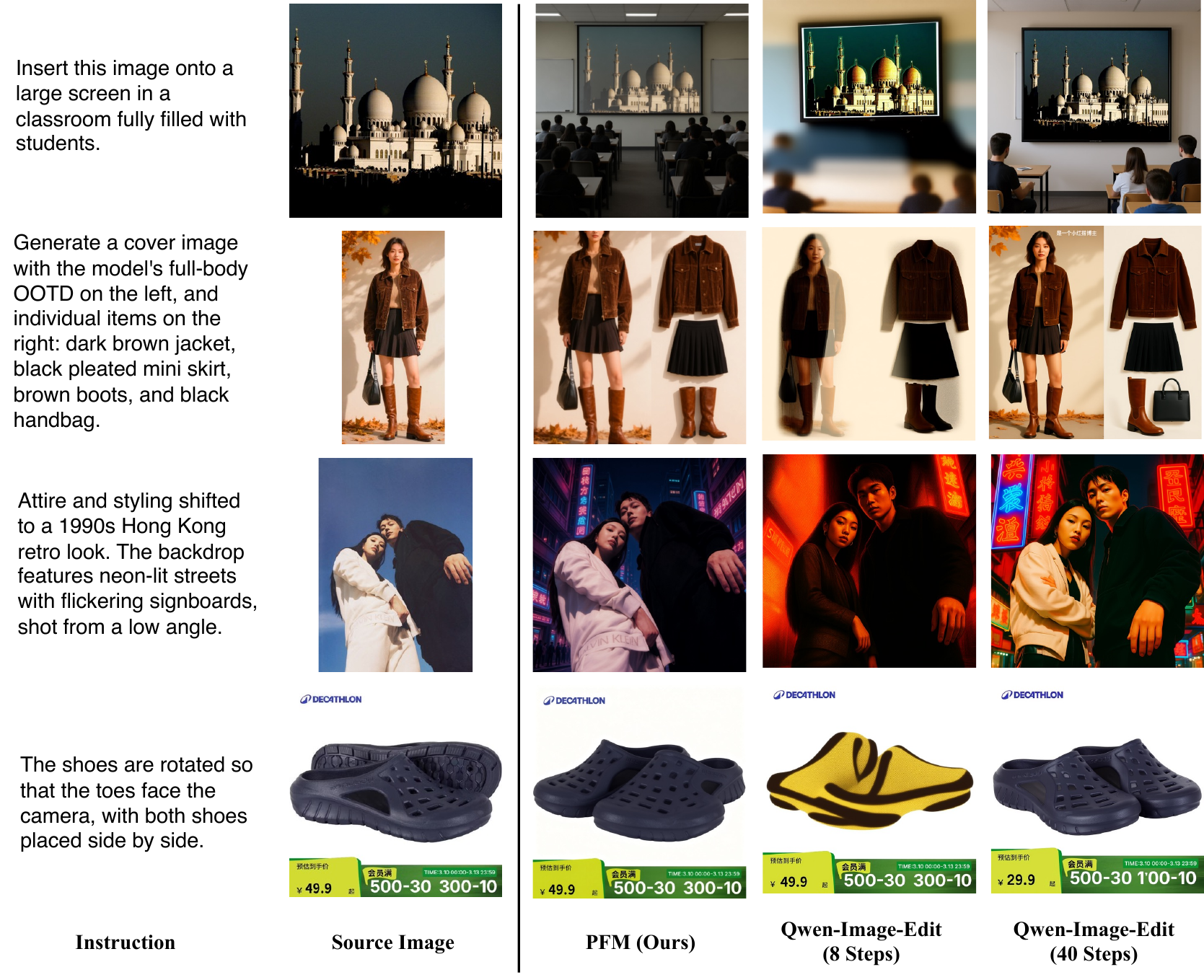}
\caption{\textbf{Qualitative results of image editing.} All images are generated with the same prompts. The Qwen-Image-Edit baselines use a CFG scale of 4.0, whereas PFM does not use CFG at inference. PFM closely matches the 40-step baseline with only 8 sampling steps.}
\label{fig:image_edit}
\end{figure}

\begin{table}[b]
\centering
\caption{\textbf{Image editing results on MagicBrush.} PFM is applied to Qwen-Image-Edit. ``$\times 2$'' denotes the additional network evaluation introduced by classifier-free guidance (CFG).}
\label{tab:magicbrush}
\setlength{\tabcolsep}{6.5pt}
\renewcommand{\arraystretch}{1.25}
\begin{tabular}{lccccccc}
\toprule
Method & NFE & CFG & CLIP-T $\uparrow$ & CLIP-I $\uparrow$ & DINO $\uparrow$ & L1 $\downarrow$ & L2 $\downarrow$ \\
\midrule
Qwen-Image-Edit & $40\times 2$ & $4.0$ & $\mathbf{0.3129}$ & $0.8957$ & $0.8312$ & $0.1060$ & $0.0428$ \\
Qwen-Image-Edit & $8\times 2$ & $4.0$ & $0.3073$ & $0.8714$ & $0.7968$ & $0.1394$ & $0.0526$ \\
PFM (ours) & $8$ & $1.0$ & $0.3026$ & $\mathbf{0.9402}$ & $\mathbf{0.9187}$ & $\mathbf{0.0535}$ & $\mathbf{0.0156}$ \\
\bottomrule
\end{tabular}
\end{table}

\section{Experiments}
\subsection{Experimental Setting}
We evaluate PFM across various tasks, including text-to-image generation, image editing, and text-to-video generation. For text-to-image generation, we apply PFM to SD3-Medium~\citep{esser2024scaling} and train
on our in-house dataset. We evaluate on COCO 2014 val~\citep{lin2014microsoft} using FID~\citep{heusel2017gans}, CLIP Score~\citep{radford2021learning}, and HPSv3~\citep{ma2025hpsv3}, which measure image
fidelity, text-image alignment, and aesthetic quality, respectively.
For image editing, we apply PFM to Qwen-Image-Edit~\citep{wu2025qwenimage}, train on our in-house dataset, and evaluate on the MagicBrush~\citep{zhang2023magicbrush} benchmark.
For text-to-video generation, we apply PFM to the Wan-2.1 1.3B model, and evaluate on the VBench~\citep{huang2024vbench} benchmark.
Detailed training settings and hyper-parameters are provided in the Appendix.

\subsection{Main Results}

\textbf{Text-to-image generation.} For text-to-image generation, we compare our method with two distillation approaches: LCM \citep{luo2023latent} and DMD2 \citep{yin2024dmd2}. All methods are trained for the same 1,000 optimization steps with the same data on SD3-Medium. As shown in Table~\ref{tab:t2i_results}, our method achieves state-of-the-art results on CLIP score, HPSv3 score and FID in 8 steps, demonstrating that simple perceptual guidance can effectively improve both semantic alignment and image quality. We further provide visual comparisons in Figure~\ref{fig:exp_sd3}. We can see that LCM produces blurrier images with more artifacts than our method, while DMD2 tends to generate images with overly high saturation and contrast.

\begin{figure}[t] \centering \includegraphics[width=\linewidth]{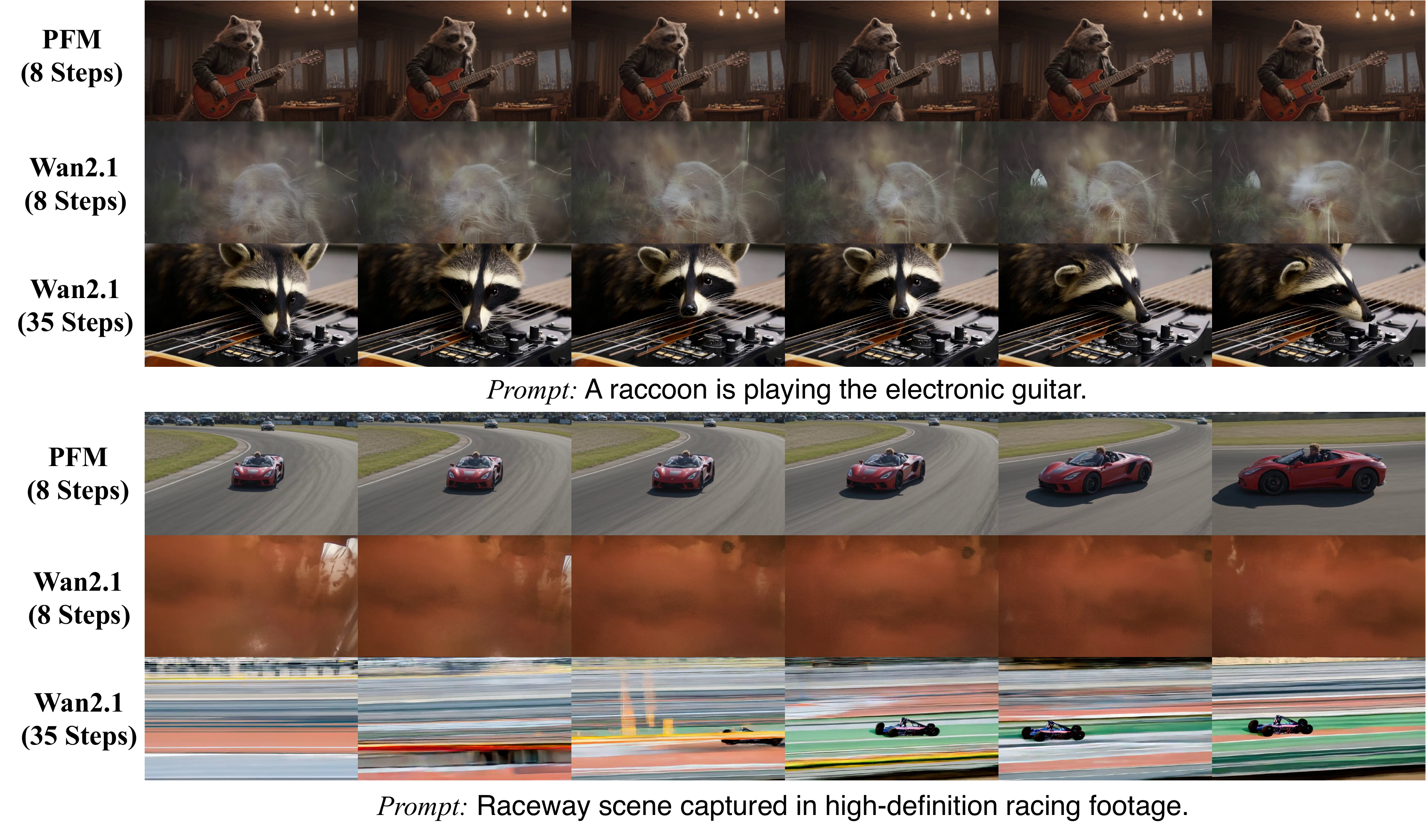} 
\caption{\textbf{Qualitative results of text-to-video generation.}} \label{fig:video_generation}
\end{figure}

\begin{table}[t]
      \centering
      \caption{\textbf{VBench evaluation results.} PFM is compared against the Wan2.1-1.3B baselines. Bold indicates the best result.}
      \label{tab:vbench}

      \setlength{\tabcolsep}{3.8pt}
      \renewcommand{\arraystretch}{1.25}
      \begin{tabular}{lcccccccccc}
      \toprule
      Method & NFE & \makecell{Object\\Class} & \makecell{Multi.\\Objects} & \makecell{Human\\Action} & Color & \makecell{Spatial\\Rel.} & Scene & \makecell{Appear.\\Style} & \makecell{Temporal\\Style} & \makecell{Overall\\Cons.} \\
      \midrule
      Wan 2.1 & 35$\times$2 & 0.786 & \textbf{0.592} & 0.810 & \textbf{0.912} & \textbf{0.646} & 0.283 & 0.222 & \textbf{0.219} & 0.219 \\
      Wan 2.1 & 8$\times$2 & 0.763 & 0.383 & 0.800 & 0.679 & 0.351 & 0.260 & \textbf{0.226} & 0.186 & 0.222 \\
      PFM & 8 & \textbf{0.820} & 0.562 & \textbf{0.910} & 0.789 & 0.620 & \textbf{0.356} & 0.198 & 0.214 & \textbf{0.246} \\
      \bottomrule
      \end{tabular}

      \vspace{4pt}

      \setlength{\tabcolsep}{2.2pt}
      \renewcommand{\arraystretch}{1.25}
      \begin{tabular}{lcccccccc|ccc}
      \toprule
      Method & NFE & \makecell{Subject\\Cons.} & \makecell{BG\\Cons.} & \makecell{Temp.\\Flicker} & \makecell{Motion\\Smooth.} & \makecell{Dynamic\\Degree} & \makecell{Aesth.\\Quality} & \makecell{Imaging\\Quality} & \makecell{Quality\\Score} & \makecell{Sem.\\Score} & \makecell{Total\\Score} \\
      \midrule
      Wan 2.1 & 35$\times$2 & 0.962 & 0.960 & 0.979 & 0.987 & 0.379 & 0.535 & \textbf{0.658} & 0.799 & 0.675 & 0.774 \\
      Wan 2.1 & 8$\times$2 & 0.918 & 0.964 & 0.994 & 0.976 & \textbf{0.514} & 0.550 & 0.465 & 0.775 & 0.578 & 0.735 \\
      PFM & 8 & \textbf{0.969} & \textbf{0.972} & \textbf{0.996} & \textbf{0.995} & 0.319 & \textbf{0.651} & 0.618 & \textbf{0.820} & \textbf{0.677} & \textbf{0.792} \\
      \bottomrule
      \end{tabular}

  \end{table}

\textbf{Image editing.}
We apply PFM to Qwen-Image-Edit and evaluate its effectiveness on image editing tasks using the MagicBrush benchmark~\citep{zhang2023magicbrush}. 
We use VGG~\citep{simonyan2014very} and DINO~\citep{caron2021emerging} as perceptual feature extractors and train the model using only the PFM objective, without CFG baking.
We compare PFM against the original Qwen-Image-Edit model with 40-step Euler sampling, as well as its 8-step consistency sampling variant with CFG scale of 4.0.

As shown in Table~\ref{tab:magicbrush}, reducing the sampling steps from 40 to 8 leads to a noticeable degradation in performance. 
In contrast, PFM significantly outperforms the 8-step baseline while requiring no classifier-free guidance at inference time.
Remarkably, PFM even surpasses the 40-step original model on CLIP-I, DINO, L1, and L2 metrics, while achieving a comparable CLIP-T score with a drop of 0.01.
These results demonstrate that PFM enables higher-quality edits with substantially fewer sampling steps, confirming its effectiveness for few-step image editing. Figure~\ref{fig:image_edit} shows qualitative comparisons of the image editing results. The 8-step baseline exhibits noticeable blurriness and artifacts, whereas PFM produces high-fidelity edits that faithfully follow the instructions, closely matching the 40-step original model.

\begin{figure}[t] \centering \includegraphics[width=\linewidth]{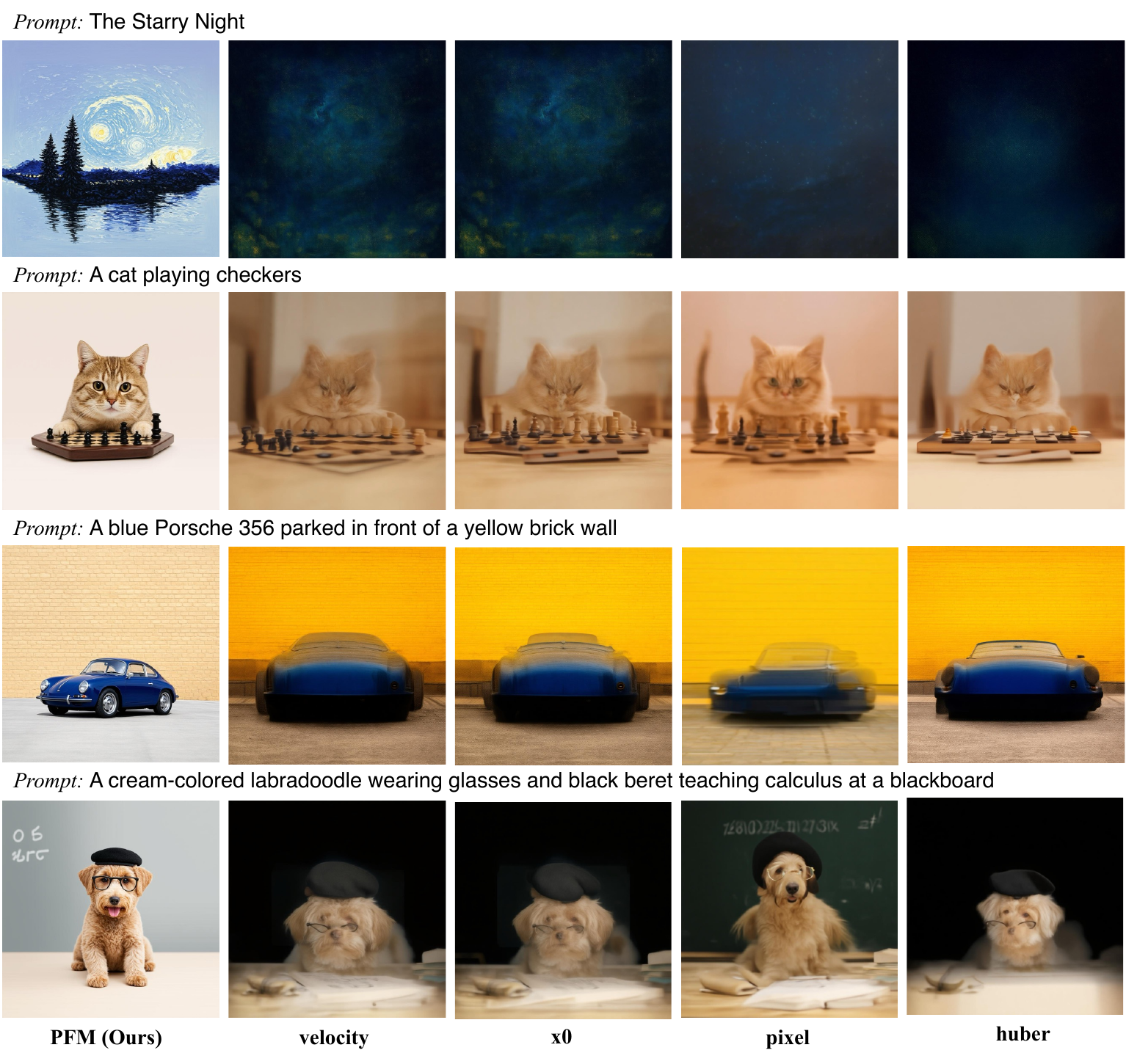} 
\caption{\textbf{Ablation of supervision spaces.} All models are fine-tuned from SD3-Medium using the same training data and the same number of training iterations, and are evaluated with 8 sampling steps. PFM is evaluated without classifier-free guidance (CFG), while all other models are evaluated with a CFG scale of $4.0$.} \label{fig:abl_supervision_space} 
\end{figure}

\textbf{Text-to-video generation.}
For text-to-video generation, we apply PFM to Wan2.1-1.3B \citep{wan2025wan} and evaluate using VBench~\citep{huang2024vbench}. We train with InternVideo2~\citep{wang2024internvideo2} as the perceptual model for 4000 iterations without CFG baking.
Empirically, we find that using a video-based foundation model such as InternVid encourages greater motion dynamics in the generated videos compared to image-based foundation models like DINOv2~\citep{oquab2023dinov2}.
We compare PFM against Wan2.1-1.3B baselines sampled with 35-step and 8-step at CFG scale of 5.0.
As shown in Table~\ref{tab:vbench}, PFM significantly surpasses the 8-step counterpart and achieves comparable, and even slightly better results than the 35-step baseline, demonstrating the effectiveness of PFM. The qualitative comparison is presented in Figure~\ref{fig:video_generation}, the 8-step
baseline exhibits significant blurriness, whereas PFM produces high-fidelity results matching the 35-step baseline.

  \begin{table}[t]
    \renewcommand{\arraystretch}{1.25}
    \centering
    \begin{minipage}{0.38\linewidth}
      \setlength{\tabcolsep}{8pt}
      \centering
      \caption{Comparison of vision backbones as feature extractors.}
      \label{tab:backbone_comparison}
      \begin{tabular}{lcc}
      \toprule
      Backbone & CLIP & HPSv3 \\
      \midrule
      VGG      & 29.5011          & 5.0642  \\
      DINO     & 29.7590          & 7.4574  \\
      ConvNeXt & \textbf{31.5326} & \textbf{10.1701} \\
      SigLIP   & 29.2188          & 5.2237  \\
      RandViT  & 20.9400          & -9.4900 \\
      \bottomrule
      \end{tabular}
    \end{minipage}
    \hfill
    \begin{minipage}{0.6\linewidth}
      \setlength{\tabcolsep}{6pt}
      \centering
      \caption{Effect of classifier-free guidance (CFG) baking and train/val CFG scales. We train with VGG+DINOv2 as perceptual models for 500 steps. $1.0$ indicates CFG is not used during inference.}
      \label{tab:cfg_baking}
      \begin{tabular}{lcccc}
      \toprule
      CFG Baking & Train Scale & Val Scale & CLIP & HPSv3 \\
      \midrule
      --          & -- & 1.0 & 30.38 & 7.41 \\
      --          & -- & 1.5 & 31.85 & 9.27 \\
      --          & -- & 2.0 & \textbf{32.16} & \underline{9.70} \\
      \midrule
      Pred side   & 1.5 & 1.0 & 31.17 & 9.17 \\
      Pred side   & 2.0 & 1.0 & 31.57 & \textbf{9.78} \\
      Pred side   & 2.5 & 1.0 & \underline{31.96} & 9.40 \\
      \midrule
      Target side & 1.5 & 1.0 & 30.13 & 7.34 \\
      Target side & 2.0 & 1.0 & 30.44 & 8.09 \\
      Target side & 2.5 & 1.0 & 30.70 & 8.62 \\
      \bottomrule
      \end{tabular}
    \end{minipage}
  \end{table}

\begin{figure}[t] \centering \includegraphics[width=\linewidth]{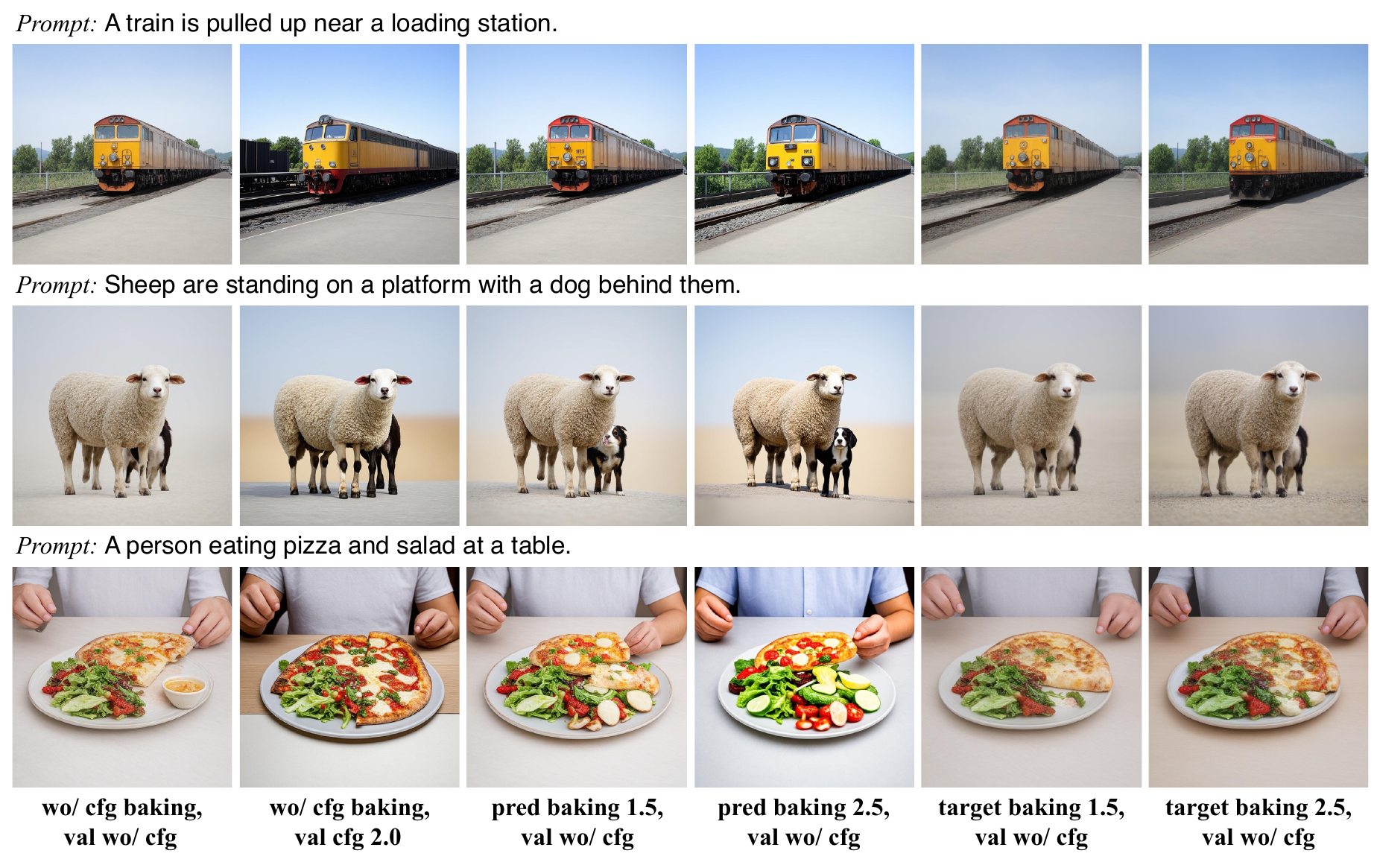} 
\caption{\textbf{Ablation of classifier-free guidance.} PFM without CFG baking generates images with good fidelity, and prediction-side CFG baking yields results similar to inference-time CFG.}
\label{fig:abl_cfg}
\end{figure}

\subsection{Ablation Studies}

\label{sec:abl_space}
\textbf{Supervision on other spaces.} 
We compare PFM against several alternative supervision strategies, including (i) regression on velocity in the VAE latent space (standard flow matching), (ii) regression on $x_0$ in the VAE latent space, (iii) regression on decoded images in pixel space, and (iv) Huber loss on $x_0$ in the VAE latent space. 
All models are trained from SD3-Medium and evaluated on PartiPrompts with $8$-step sampling. 
As shown in Figure~\ref{fig:abl_supervision_space}, only perceptual supervision produces sharp and high-quality results, while all alternative supervision spaces yield noticeably blurrier outputs.
These results suggest that the supervision space plays a critical role in few-step generation, and that perceptual features provide a substantially more effective training signal than latent- or pixel-space regression.

\begin{figure}[t] \centering \includegraphics[width=\linewidth]{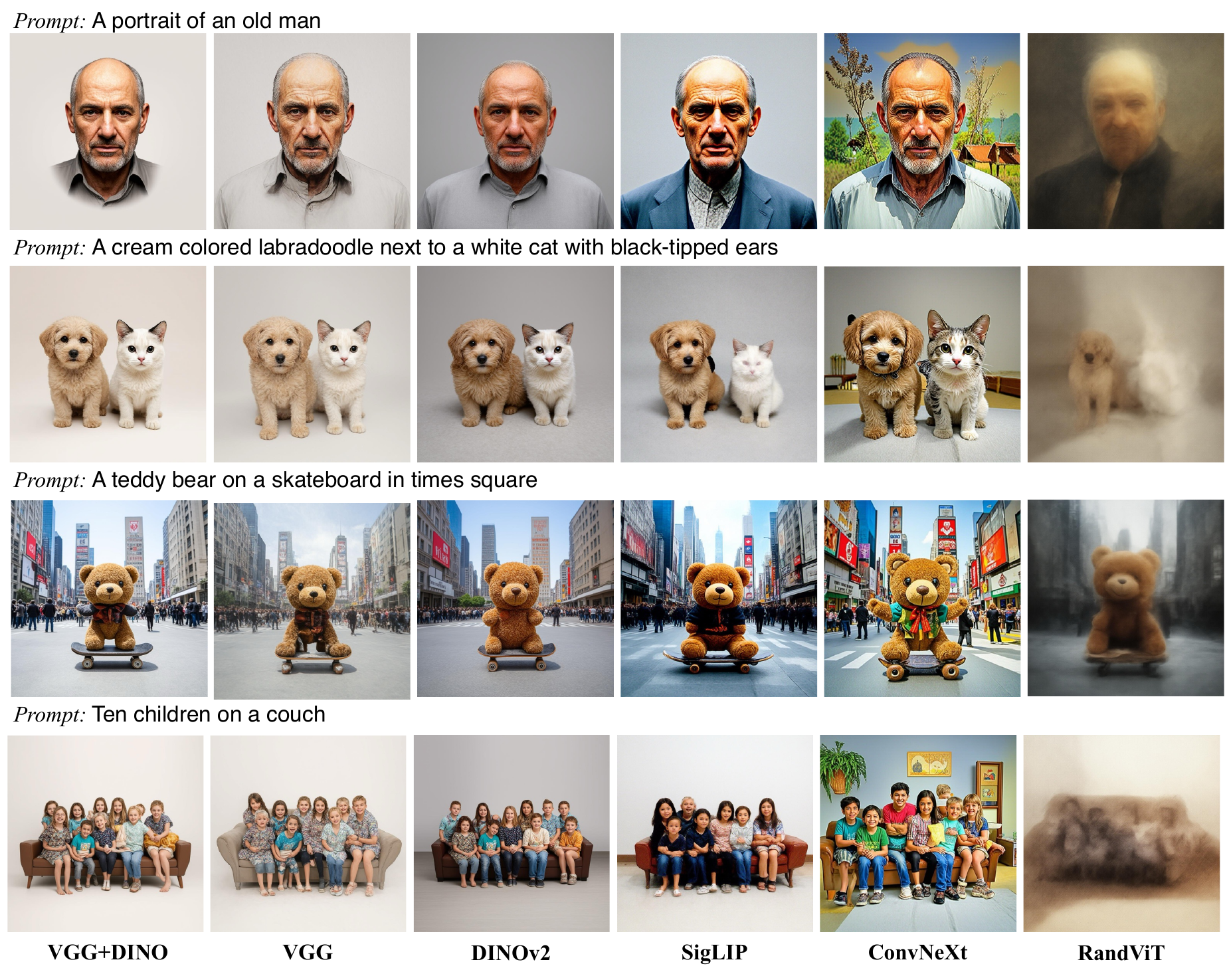} 
\caption{\textbf{Ablation of different perceptual models.} All perceptual models yield clear results in 8 steps, while a randomly initialized ViT (RandViT) produces blurry results, similar to standard flow
matching.}
  \label{fig:abl_perceptual_model}
\end{figure}

\label{sec:abl_pm}
\textbf{Perceptual models.}
We compare several perceptual models: VGG, DINOv2, SigLIP~\citep{zhai2023sigmoid}, ConvNeXt~\citep{liu2022convnet}, a randomly initialized ViT (RandViT), and the combination of VGG+DINOv2.
As shown in Figure~\ref{fig:abl_perceptual_model}, all pretrained models produce sharp and visually coherent results, whereas RandViT yields noticeably blurrier outputs, confirming that learned visual representations are essential for effective perceptual supervision.
Among pretrained models, those operating at higher resolutions (DINOv2, SigLIP, ConvNeXt) preserve more fine-grained details — particularly in background regions — than VGG, which is limited to $224 \times 224$ inputs.
We also observe that different perceptual models lead to distinct visual characteristics, and that combining perceptual models yields better performance. We further evaluate the text-to-image metrics on COCO 2014
val using SD3-Medium trained with different perceptual models in Table~\ref{tab:backbone_comparison}. Surprisingly, the performance is closely related to the off-manifold distance we measured in
Table~\ref{tab:why_perceptual_spaces}.

\label{sec:abl_cfg}
\textbf{Classifier-free guidance.}
As shown in Figure~\ref{fig:abl_cfg}, PFM trained with only the perceptual objective already generates high-quality samples without classifier-free guidance, suggesting that the guidance effect is implicitly learned when using perceptual objective.
As shown in Table~\ref{tab:cfg_baking}, the model remains compatible with standard CFG at inference time, where relatively small guidance scales suffice to achieve strong results. 
When applying the proposed CFG baking strategies (Section~\ref{sec:cfg}), and the model trained with CFG baking yields results closely matching inference-time CFG while requiring only a single forward pass during sampling.

\begin{figure}[t] \centering \includegraphics[width=\linewidth]{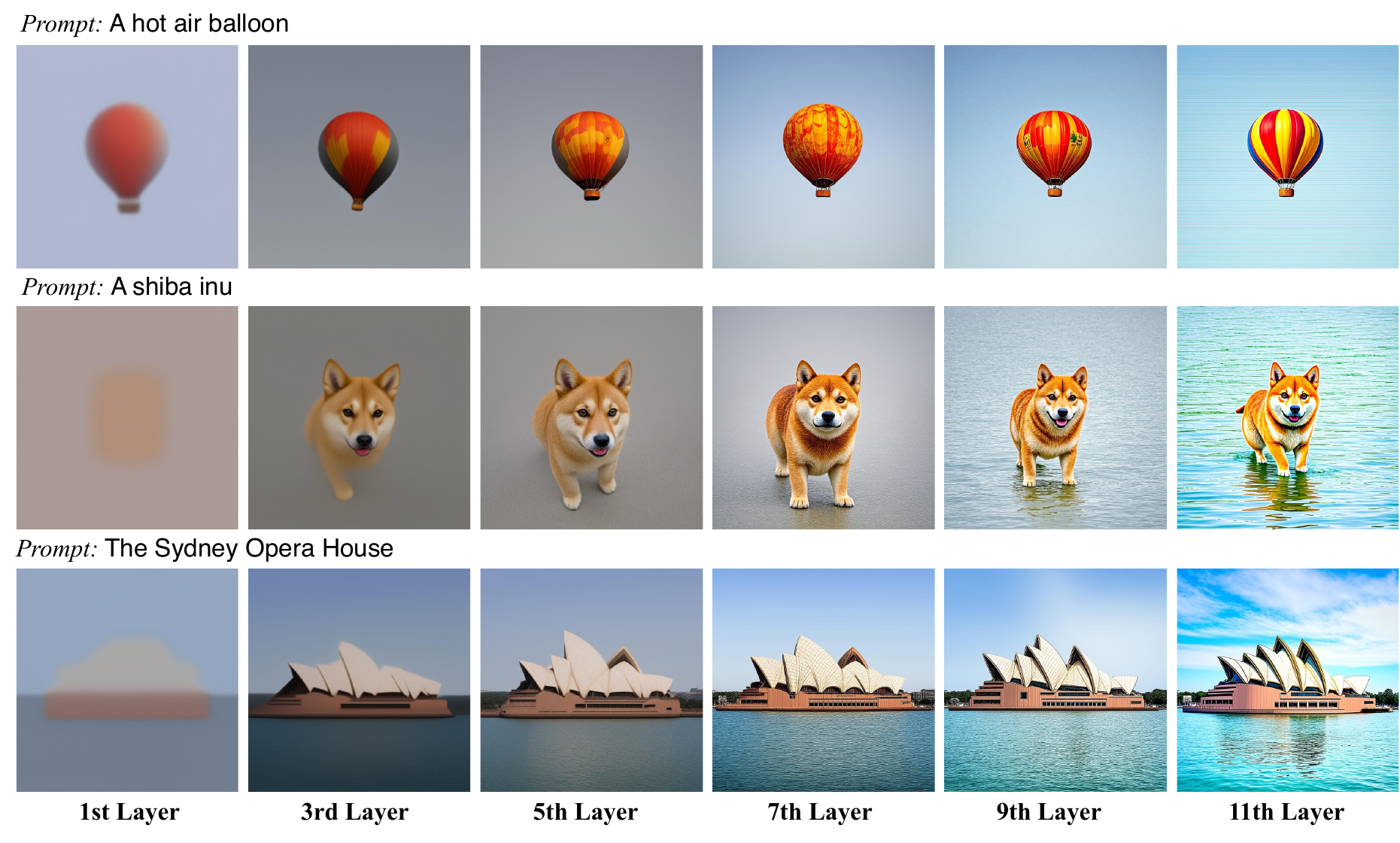} 
\caption{\textbf{Ablation on layer depth in the perceptual model.} We compute the perceptual loss using features from different layers of DINOv2. Shallow layers, which remain closer to pixel-level representations, produce blurrier results, while the deepest layers tend to collapse to fixed modes.}
\label{fig:abl_depth}
\end{figure}

\begin{figure}[t] \centering \includegraphics[width=\linewidth]{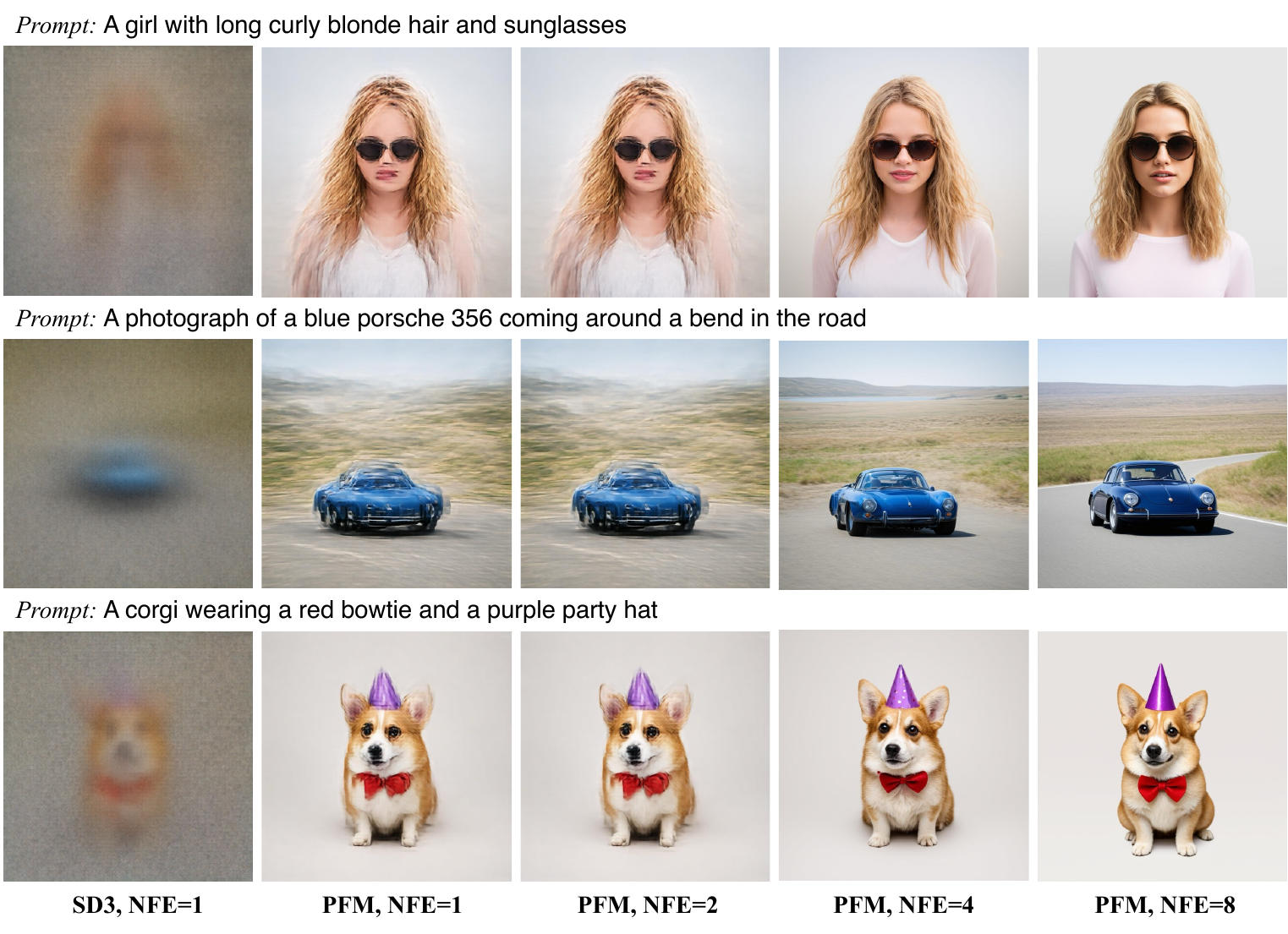} 
\caption{\textbf{Ablation of inference steps.} PFM produces high-fidelity results at 4 and 8 NFE and remains sharp even at 1 NFE, whereas the SD3 baseline is blurry at 1 NFE.}
\label{fig:abl_nfe}
\end{figure}

\label{sec:abl_steps}
\textbf{Inference steps.} We evaluate PFM with prediction-side CFG baking on PartiPrompts using 1, 2, 4, and 8 sampling steps. 
As shown in Figure~\ref{fig:abl_nfe}, generation quality improves consistently with more steps. 
At 1--2 steps, the model produces recognizable but noticeably blurry images; at 4--8 steps, the results achieve strong visual quality with most fine details preserved.
We leave extending PFM to the 1--2 step as future work.

\label{sec:abl_depth}
\textbf{Layer depth in the perceptual model.} We ablate which DINOv2 layer the perceptual
loss is applied to. 
As shown in Figure~\ref{fig:abl_depth}, supervising on shallow layers yields substantially blurrier results than
deeper layers, consistent with our finding in {Section~\ref{sec:why_perceptual}}
that a randomly initialized network fails: shallow features remain nearly pixel-isometric,
whereas deeper layers encode the learned semantic structure that makes the supervision metric
non-trivial. However, supervising solely on the deepest layers biases the model toward
abstract features invariant to fine appearance, degrading pixel fidelity; we therefore average
features across multiple layers.

\section{Conclusion}
We introduced Perceptual Flow Matching (PFM), a simple approach for few-step generation that changes only the supervision space of flow matching. By supervising decoded clean predictions with pretrained perceptual features instead of regressing velocities in the VAE latent space, PFM encourages sharper and more visually coherent predictions under large sampling steps.
PFM preserves the standard flow-matching framework and requires no teacher model, auxiliary score network, or specialized distillation pipeline. Across image generation, image editing, and video generation, it enables high-quality sampling in 4--8 steps. Our ablations show that the supervision geometry is critical: pretrained perceptual representations substantially improve few-step generation, while latent- or pixel-space regression remains prone to blurry predictions.

PFM still has several limitations. Its performance degrades in extremely low-step regimes, especially with fewer than two sampling steps. In addition, although our results suggest that perceptual spaces provide a
more favorable geometry for few-step prediction, which perceptual space is optimal remains unknown. We leave stronger one-step generation and a deeper characterization of perceptual supervision spaces to future
work

\clearpage

\bibliographystyle{plainnat}
\bibliography{references}

\end{document}